\documentclass{article} 
\usepackage{nips15submit_e,times}
\usepackage{url}
\usepackage{latexsym}
\usepackage{graphicx}
\usepackage[toc,page]{appendix}

\usepackage{pdflscape}

\usepackage{amsmath}
\usepackage{amsfonts}

\usepackage{slashbox}
\newenvironment{myfont}{\fontfamily{pcr}\selectfont}{\par}
\DeclareTextFontCommand{\textmyfont}{\myfont}

\usepackage{bm}
\newcommand{\mat}[1]{\boldsymbol{\mathbf{#1}}}
\newcommand{\argmin}{\operatornamewithlimits{argmin}}

\newcommand{\model}[1]{{\small{#1}}}
\newcommand{\bb}[1]{\mathbb{#1}}

\newcounter{tbsnr}
\newenvironment{tbs}
{\addtocounter{tbsnr}{1}\par\bigskip \noindent\fbox{\thetbsnr}
\hspace*{\fill}\begin{minipage}{10cm}\tt}
{\end{minipage}\hspace*{\fill}\bigskip}

\title{Unveiling the Dreams of Word Embeddings:\\
Towards Language-Driven Image Generation}

\author{
Angeliki Lazaridou\\
Center for Mind/Brain Sciences\\
University of Trento\\
\texttt{angeliki.lazaridou@unitn.it} \\
\And
Dat Tien Nguyen\thanks{Research carried out in Center for Mind/Brain Sciences, University of Trento} \\
Institute of Formal and Applied Linguistics\\
Charles University of Prague \\
\texttt{dtnguyen88@outlook.com}\\
\AND
Raffaella Bernardi \\
DISI and Center for Mind/Brain Sciences\\
University of Trento \\
\texttt{raffaella.bernardi@unitn.it}\\
\And
Marco Baroni \\
Center for Mind/Brain Sciences\\
University of Trento\\
\texttt{marco.baroni@unitn.it}}

\nipsfinalcopy 

\begin{document}
\maketitle
\begin{abstract}
  We introduce \emph{language-driven image generation}, the task of
  generating an image visualizing the semantic contents of a word
  embedding, e.g., given the word embedding of \emph{grasshopper}, we
  generate a natural image of a grasshopper.  We implement a simple
  method based on two mapping functions. The first takes as input a
  word embedding (as produced, e.g., by the word2vec toolkit) and maps
  it onto a high-level visual space (e.g., the space defined by one of
  the top layers of a Convolutional Neural Network). The second
  function maps this abstract visual representation to pixel space, in
  order to generate the target image. Several user studies suggest
  that the current system produces images that capture general visual
  properties of the concepts encoded in the word embedding, such as
  color or typical environment, and are sufficient to discriminate
  between general categories of objects.
\end{abstract}

\section{Introduction}
\label{sec:introduction}

\emph{Imagination}, creating new images in the mind, is a fundamental
capability of humans, studies of which date back to Plato's ideas
about memory and perception. Through imagery, we form \emph{mental
  images}, picture-like representations in our mind, that encode and
extend our perceptual and linguistic experience of the world.  Recent
work in neuroscience attempts to generate reconstructions of these
mental images, as encoded in vector-based representations of fMRI
patterns~\cite{Nishimoto:etal:2011}.  In this work, we take the first
steps towards implementing the same paradigm in a computational setup,
by generating images that reflect the imagery of distributed
\emph{word representations}.

We introduce \emph{language-driven image generation}, the task of
visualizing the contents of a linguistic message, as encoded in word
embeddings, by generating a real image.  Language-driven image
generation can serve as evaluation tool providing intuitive
visualization of what computational representations of word meaning
encode. More ambitiously, effective language-driven image generation
could complement image search and retrieval, producing images for
words that are not associated to images in a certain collection,
either for sparsity, or due to their inherent properties (e.g.,
artists and psychologists might be interested in images of abstract or
novel words). In this work, we focus on generating images for
distributed representations encoding the meaning of \emph{single}
words. However, given recent advances in compositional distributed
semantics~\cite{Socher:etal:2013b} that produce embeddings for
arbitrarily long linguistic units, we also see our contribution as the
first step towards generating images depicting the meaning of phrases
(e.g., \emph{blue car}) and sentences.  After all, language-driven
image generation can be seen as the symmetric goal of recent research
(e.g., \cite{Karpaty:FeiFei:2014,Kiros:etal:2014b}) that introduced
effective methods to generate linguistic descriptions of the contents
of a given image.

To perform language-driven image generation, we combine various recent strands
of research.  Tools such as word2vec \cite{Mikolov:etal:2013b} and Glove
\cite{Pennington:etal:2014} have been shown to produce extremely high-quality
vector-based \emph{word embeddings}. At the same time, in computer vision,
images are effectively represented by vectors of abstract visual features, such
as those extracted by Convolutional Neural Networks (CNNs)
\cite{Krizhevsky:etal:2012}. Consequently, the problem of translating between
linguistic and visual representations has been coached in terms of learning a
\emph{cross-modal mapping} function between vector spaces
\cite{Frome:etal:2013,Socher:etal:2013a}. Finally, recent
work in computer vision, motivated by the desire to achieve a better
understanding of what the layers of CNNs and other deep architectures have
really learned, has proposed \emph{feature inversion} techniques that map a
representation in abstract visual feature space (e.g., from the top layer of a
CNN) back onto pixel space, to produce a real image
\cite{Zeiler:Fergus:2014,Mahendran:Vedaldi:2015}.

Our language-driven image generation system takes a word embedding as
input (e.g., the word2vec vector for \emph{grasshopper}), projects it
with a cross-modal function onto visual space (e.g., onto a
representation in the space defined by a CNN layer), and then applies
feature inversion to it (using the method HOGgles method of
\cite{Vondrick:etal:2014}) to generate an actual image (cell A18 in
Figure \ref{fig:summary}). We test our system in a rigorous
\emph{zero-shot} setup, in which words and images of tested concepts
are neither used to train cross-modal mapping, nor employed to induce
the feature inversion function. So, for example, our system mapped
\emph{grasshopper} onto visual and then pixel space without having
ever been exposed to grasshopper pictures.

Figure \ref{fig:summary} illustrates our results (``answer key'' for
the figure provided as supplementary material). While it is difficult
to discriminate among similar objects based on these images, the
figure shows that our language-driven image generation method already
captures the broad gist of different domains (food looks like food,
animals are blobs in a natural environment, and so on).

\begin{figure*}[t]
\begin{centering}
\includegraphics[scale=0.55]{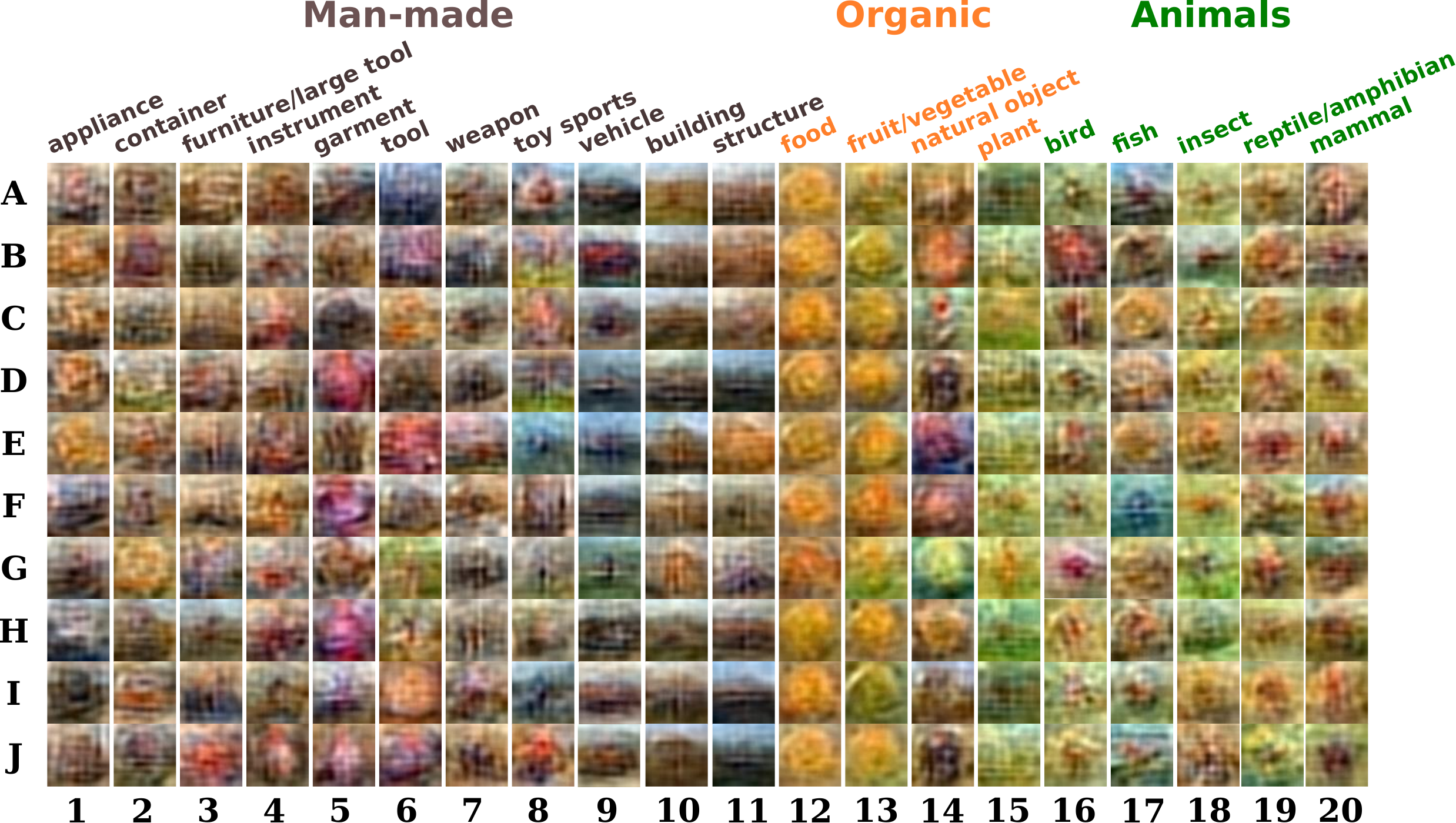}
\end{centering}
\caption{Generated images of 10 concepts per category for 20 basic
  categories, grouped my macro-category. See supplementary materials
  for the answer key.\label{fig:summary}}
\end{figure*}

\section{Language-driven image generation}
\label{sec:pipeline} 


\subsection{From word to visual vectors}
\label{sec:from-word-to-visual-vectors} Up to now, feature inversion algorithms
\cite{Mahendran:Vedaldi:2015,Vondrick:etal:2014,Zeiler:Fergus:2014} have been
applied to \emph{visual} representations directly extracted from images (hence
the ``inversion'' name).  We aim instead at generating an image conveying the
semantics of a concept as encoded in a \emph{word} representation. Thus, we
need a way to ``translate'' the word representation into a visual
representation, i.e., a representation laying on the visual space that conveys
the corresponding \emph{visual} semantics of the word.

Cross-modal mapping has been first introduced in the context of zero-shot
learning as a way to address the manual annotation bottleneck in domains where
other vector-based representations (e.g., images or brain signals) must be
associated to word labels~\cite{Mitchell:etal:2008,Socher:etal:2013a}. This is
achieved by using training data to learn a mapping function from vectors in the
domain of interest to vector representations of word labels.  In our case, we
are interested in the general ability of cross-modal mapping to
\emph{translate} a representation between different spaces, and specifically
from a word to a visual feature space.

The mapping is performed by inducing a function $f: \mathbb{R}^{d_{1}}\to
\mathbb{R}^{d_2}$ from data points $(w_i, v_i)$, where $w_i \in
\mathbb{R}^{d_1}$ is a word representation and $v_i \in \mathbb{R}^{d_2}$ the
corresponding visual representation.  The mapping function can then be applied
to any given word vector $w_j$ to obtain its projection $\hat{v}_j = f(w_j)$
onto visual space.  Following previous work
\cite{Mitchell:etal:2008,Frome:etal:2013}, we assume that the mapping is
linear. To estimate its parameters $\mat{M} \in \bb{R}^{d1 \times d2}$, given
word vectors $\mat{W}$ paired with visual vectors $\mat{V}$, we use
Elastic-Net-penalized least squares regression, that linearly combines the L1
and L2 weight penalties of Lasso and Ridge regularization:
\begin{equation}
\hat{\mat{M}} = \argmin_{\mat{M} \in \bb{R}^{d_1 \times d_2}}\|\mat{W}\mat{M}-\mat{V}\|_F + \lambda_1\|\mat{M}\|_1 +\lambda_2\|\mat{M}\|_F
\label{eq:mapping}
\end{equation}

By modifying the weights of the L1 and L2 penalties, $\lambda_1$ and
$\lambda_2$, we can derive different regression methods.  Specifically, we
experiment with \model{plain} regression ($\lambda_1=0$, $\lambda_2=0$),
\model{ridge} regression ($\lambda_1=0$, $\lambda_2\neq0$), \model{lasso}
regression ($\lambda_1\neq0$ and $\lambda_2=0$) and \model{symmetric elastic
net} ($\lambda_1=\lambda_2$, $\lambda_1\neq0$).

\subsection{From visual vectors to images} 
\label{sec:from-visual-vectors-to-images}
Convolutional Neural Networks have recently surpassed human performance on
object recognition~\cite{Russakovsky:etal:2014}. Nevertheless, these models
exhibit ``intriguing properties'', that are somewhat surprising given their
state-of-the-art performance~\cite{Szegedy:etal:2013}, prompting an effort to
reach a deeper understanding of how they really work.  Given that these models
consist of millions of parameters, there is ongoing research on feature
inversion of different CNN layers to attain an intuitive visualization of what
each of them learned.

Several methods have been proposed for inverting CNN visual features,
however, the exact nature of the task imposes certain constraints on
the inversion method.  For example, the original work of Zeiler and
Fergus~\cite{Zeiler:Fergus:2014} cannot be straightforwardly adapted
to our task of generating images from word embeddings, since their
\emph{DeConvNet} method requires information related to the
activations of the network in several layers. In this work, we adopt
the framework of Vondrick et al.~\cite{Vondrick:etal:2014} that casts
the problem of inversion as \emph{paired dictionary
  learning}.\footnote{Originally, the HOGgles method
  of~\cite{Vondrick:etal:2014} was introduced for visualizing HOG
  features. However, the method does not make feature-specific
  assumptions and it has also recently been used to invert CNN
  features~\cite{Vondrick:etal:2014b}.}

Specifically, given an image $x_0 \in \bb{R}^D$ and its visual
representation $y=\phi(x_0) \in \bb{R}^d$, the goal is to find an image
$x^{\ast}$  that minimizes the reconstruction error:
\begin{equation}
x^{\ast} = \argmin_{x \in \bb{R}^D} \| \phi(x) - y \|_{2}^{2}
\label{eq:objective}
\end{equation}
Given that there are no guarantees regarding the convexity of $\phi$,
both images and visual representations are approximated by paired,
over-complete bases, $U \in \bb{R}^{D \times K}$ and $V \in \bb{R}^{d
  \times K}$, respectively.  Enforcing $U$ and $V$ to have paired
representations through shared coefficients $\alpha \in \mathbb{R}^K$, i.e., 
 $x_0 = U\alpha$ and $y=V\alpha$, allows the
feature inversion to be done by estimating such coefficients $\alpha$ that
minimize the reconstruction error. Practically, the algorithm
proceeds by finding $U$, $V$ and $\alpha$ through a standard sparse
coding method. For learning the parameters, the algorithm is presented with training
data of the form $(x_i,y_i)$, where $x_i$ is an image patch and $y_i$
the corresponding visual vector associated with that
patch. 

\section{Experimental Setup}

\subsection{Materials}

\paragraph{Dreamed Concepts} We refer to the words we generate images
for as \emph{dreamed} concepts. The dreamed word set comes from the
concepts studied by McRae et al.~\cite{McRae:etal:2005}, in the
context of property norm generation. This set contains 541 base-level
\emph{concrete} concepts (e.g., cat, apple, car etc.) that span across
20 general and broad categories (e.g., animal, fruit/vegetable,
vehicle etc).  For the purposes of the current experiments, 69 McRae
concepts were excluded (either because of high ambiguity or for
technical reasons), resulting in 472 dreamed words we test on.

\paragraph{Seen Concepts} We refer to the set of words associated to
real pictures that are used for training purposes as \emph{seen}
concepts. The real picture set contains approximately 480K images
extracted from ImageNet \cite{Deng:etal:2009} representing 5K distinct
concepts. The seen concepts are used for training the cross-modal
mapping. Importantly, the dreamed and seen concept sets do emph{not} overlap.

\paragraph{Word Representations} For all seen and dreamed concepts, we
build 300-dimensional word vectors with the word2vec
toolkit,\footnote{\url{https://code.google.com/p/word2vec/}} choosing
the CBOW method.\footnote{Other hyperparameters, adopted without
  tuning, include a context window size of 5 words to either side of
  the target, setting the sub-sampling option to 1e-05 and estimating
  the probability of target words by negative sampling, drawing 10
  samples from the noise distribution~\cite{Mikolov:etal:2013b}.}
CBOW, which learns to predict a target word from the ones surrounding
it, produces state-of-the-art results in many linguistic
tasks~\cite{Baroni:etal:2014}. Word vectors are induced from a
language corpus (e.g., Wikipedia)
of 2.8 billion words.\footnote{Corpus sources:
  \url{http://wacky.sslmit.unibo.it},
  \url{http://www.natcorp.ox.ac.uk}}

\paragraph{Visual Representations} The visual representations, for the
set of 480K seen concept images, are extracted with the pre-trained
CNN model of \cite{Krizhevsky:etal:2012} through the Caffe toolkit
\cite{Jia:etal:2014}. CNNs trained on natural images learn a hierarchy
of increasingly more abstract properties: the features in the bottom
layers resemble Gabor filters, while features in the top layers
capture more abstract properties of the dataset or tasks the CNN is
trained for (see \cite{Zeiler:Fergus:2014}) (e.g., the topmost layer
captures a distribution over training labels).  In this work, we
experiment with feature representations extracted from two levels,
\model{pool-5}, extracted from the 5th layer (6x6x256=9216
dimensions), and \model{fc-7}, extracted from the 7th layer (1x4096
dimensions). \model{pool-5} is an intermediate pooling layer that
should capture object commonalities. \model{fc-7} is a fully-connected
layer just below the topmost one, and as such it is expected to
capture high-level discriminative features of different object
classes.

Since each seen concept is associated with many images, we experiment
with two ways to derive a unique visual representation. Inspired from
categorization schemes in cognitive science~\cite{Murphy:2002}, we
will refer to them as the \emph{prototype} and \emph{exemplar}
methods. The \emph{prototype} visual vector of a concept
is constructed by averaging the visual representations (either
\model{pool-5} or \model{fc-7}) of images tagged in ImageNet with the
concept. The averaging method should smooth out noise and emphasize
invariances in images associated to a concept. On the other hand, the
constructed prototype does not correspond to an actual depiction of
the concept. The \emph{exemplar} visual vector, on the other hand, is
a single visual vector that is picked as a good representative of the
set, as it is the one with the highest average cosine similarity to
all other vectors extracted from images labeled with the same concept.

\subsection{Model selection and parameter estimation}

\paragraph{Visual feature type and concept representations} In order
to determine the optimal visual feature type (between \model{pool-5}
and \model{fc-7}) and concept representation method (between
\model{prototype} and \model{exemplar}), we set up a human study
through CrowdFlower.\footnote{\url{http://www.crowdflower.com/}} For
50 randomly chosen test concepts, we generate 4 images, each obtained
by inverting the visual vector computed by combining a feature type
with a concept representation method, e.g., for
\model{pool-5+prototype}, we generate an image by inverting the visual
vector resulting from \emph{averaging} the \model{pool-5} feature
vectors extracted from images labeled with the test concept (details
on our implementation of feature inversion below). Participants are
then asked to judge which of the 4 images is more likely to denote the
test concept. For each test concept, we collect 20 judgments. Overall,
participants showed a strong significant preference for the images
generated from inverting \model{pool-5} feature vectors (28/50), and
in particular for those that were generated from \model{pool-5} by
inverting feature vectors constructed with the \model{exemplar}
protocol (18/50).\footnote{Throughout this paper, statistical
  significance is assessed with two-tailed exact binomial tests with
  threshold $\alpha<0.05$, corrected for multiple comparisons with the
  false discovery rate method.} The following experiments were thus
carried out using the \model{pool-5+exemplar} visual space.

%

\paragraph{Cross-modal mapping} To learn the mapping $\mat{M}$ of
Equation~\ref{eq:mapping}, we use 5K training pairs ($\mathbf{w}_c,
\mathbf{v}_c$) $= \{\mathbf{w}_c \in \bb{R}^{300}, \mathbf{v}_c \in
\bb{R}^{9216}\}$, where $\mathbf{w}_c$ is the word vector and
$\mathbf{v}_c$ is the visual vector for the (seen) concept $c$, based
on \model{pool-5} features and \model{exemplar} representation.
Specifically, we estimate the weights $\mat{M}$ by training the 4
regression methods described in
Section \ref{sec:from-word-to-visual-vectors} above,
cross-validating the values of $\lambda_1$ and $\lambda_2$ on the
training data.  Model selection is performed by conducting a human
study on the language-driven image generation task. For the same test of
50 concepts as above, we obtain estimates of their visual vectors
$\hat{v}$ by mapping their word vectors into visual space through the
different mapping functions $\mat{M}$.  We then generate an image by
inverting the visual features $\hat{v}$.  Participants are again asked
to judge which of the 4 images is more likely to denote the test
concept. For each concept we collected 20 judgments. Participants
showed a preference for \model{plain} regression (9/50 significant
tests in favor of this model), which we adopt in rest of the paper.


\paragraph{Feature inversion} Training data for feature inversion (Section
\ref{sec:from-visual-vectors-to-images} above) are created by using the PASCAL
VOC 2011 dataset, that contains 15K images of 20 distinct objects. Note that
the 20 PASCAL objects are not part of our dreamed concepts, and thus the
feature inversion is performed in a zero-shot way 
(the inversion will be asked to generate an image for a concept that it has never encountered before). 
In order to increase the size of the training data, from each image we derived several image patches $x_i$
associated with different parts of the image and paired them with their
equivalent visual representations $y_i$.  Both paired dictionary learning and
feature inversion are conducted using the HOGgles
software~\cite{Vondrick:etal:2014} with default
hyperparameters.\footnote{\url{https://github.com/CSAILVision/ihog}}

\section{Experiments}
\label{sec:experiments}

Figure~\ref{fig:summary} provides a snapshot of our results; we
randomly picked 10 dreamed concepts from each of the 20 McRae
categories, and we show the image we generated for them from the
corresponding word embeddings, as described in
Section~\ref{sec:pipeline}.  We stress again that the images of
dreamed concepts were never used in any step of the pipeline, neither
to train cross-modal mapping, nor to train feature inversion, so they
are genuinely generated in a zero-shot manner, by leveraging their
linguistic associations to seen concepts.

Not surprisingly, the images we generate are not as clear as those one
would get by retrieving existing images. However, we see in the figure
that concepts belonging to different categories are clearly
distinguished, with the exception of \emph{food} and
\emph{fruit/vegetable} (columns 12 and 13), that look very much the
same (on the other hand, fruit and vegetable \emph{are} also food, and
word vectors extracted from corpora will likely emphasize this
``functional'' role of theirs).

We next present a series of user studies providing
quantitative and qualitative insights into the information that
subjects can extract from the visual properties of the generated
images.

\subsection{Experiment 1: Correct word vs.~random confounder}


The first experiment is a sanity check, evaluating whether the visual
properties in the generated images are informative enough for subjects
to guess the correct label against a random alternative.

\paragraph{Experiment description} Participants are presented with the
generated image of a dreamed concept and are asked to judge if it is
more likely to denote the correct word or a confounder randomly picked
from the seen word set.  Given that the confounder is a randomly
picked item, the task is relatively easy.  However, both confounders
and dreamed concepts are concrete, basic-level concepts, so they are
sometimes related just by chance.  Moreover, the confounders were used
to train the mapping and inversion functions, which could have
introduced a systematic bias in their favour.  We test the 472 dreamed
concepts, collecting 20 ratings for each via CrowdFlower.  Word order
is randomized both across and within trials (the same setup is used in
the following experiments, with image order also randomized).

\paragraph{Results} Participants show a consistent preference for the
correct word (dreamed concept) (median proportion of votes in favor:
75\%). Preference for the correct word is significantly different from
chance in 211/472 cases. Participants expressed a significant
preference for the confounder in 10 cases only, and in the majority of
those, dreamed concepts and their confounders shared similar
properties, e.g., \textit{cape}-\textit{tabletop} (both made of
textile), \textit{zebra}-\textit{baboon} (both mammals),
\textit{oak}-\textit{boathouse} (existing in similar natural
environments).



The experiment confirms that our method can generally capture at least
those visual properties of dreamed concepts that can distinguish them
from visually dissimilar random items.




\subsection{Experiment 2: Correct image vs.~image of similar concept}

The second experiment ascertains to what extent subjects can pick the
right generated image for a dreamed concept over a closely related
alternative.

\paragraph{Experiment description} For each dreamed concept, we pick
as confounder the closest semantic neighbor according to the
subject-based conceptual distance statistics provided by McRae et
al.~\cite{McRae:etal:2005}. In 379/472 cases, the confounder belongs
to the category of the dreamed concept; hence, distinguishing the
two concepts is quite challenging (e.g., \textit{mandarin} vs.\
\textit{pumpkin}). Participants were presented with the images
generated from the dreamed concept and the confounder, and they were
asked which of the two images is more likely to denote the dreamed
concept.

\paragraph{Results} Results and examples are provided in Table~\ref{tab:exp4}.  In the
vast majority of cases (409/472) the participants did not show a significant
preference for either the correct image or the confounder. This shows that the
current image generation pipeline does not capture, yet, fine-grained
properties that would allow within-category discrimination. Still, within the
subset of 63 cases for which subjects did express a significant preferences, we
observe a clear trend in favour of the correct image (41 vs.~22). Color and
environment seem to be the fine-grained properties that determined many of the
subjects' right or wrong choices within this subset. Of the 63 pairs, 14
involve concepts from different categories, and 49 same-category pairs. Of the
former, in 11/14 the preference was for the right image. In  2 of the 3 wrong
cases, the dreamed concept vs.~intruder pairs have similar color
(\textit{emerald} vs.\ \textit{parsley}, \textit{bowl} vs.\ \textit{dish}),
while neither concept has a typical discriminative color in the third case
(\textit{thermometer} vs.\ \textit{marble}). Even in the challenging
same-category group, 30/49 pairs display the right preference. In particular,
subjects distinguished objects that typically have different colors (e.g.,
\textit{flamingo} vs.\ \textit{partridge}), or live in different environments
(e.g., \textit{turtle} vs.\ \textit{tortoise}).  In the remaining 19
within-category cases in which the confounder was preferred, color seems again
to play a crucial role in the confusion (e.g., \textit{alligator} vs.\
\textit{crocodile}, \textit{asparagus} vs.\ \textit{spinach}).

\begin{table}[t]
\begin{center}
    \begin{tabular}{c c |c c }
\multicolumn{2}{c}{In favor of \textbf{dreamed} concept}                 & \multicolumn{2}{c}{In favor of \textbf{confounder}} \\
\multicolumn{2}{c}{8.6\% (41/472)} &\multicolumn{2}{c}{4.6\% (22/472)}\\
Same category & Different category &Same category & Different category\\\hline\hline
\includegraphics[scale=0.5]{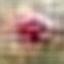}\includegraphics[scale=0.5]{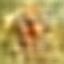} &\includegraphics[scale=0.5]{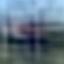}\includegraphics[scale=0.5]{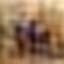} & \includegraphics[scale=0.5]{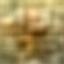}\includegraphics[scale=0.5]{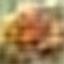} &\includegraphics[scale=0.5]{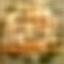}\includegraphics[scale=0.5]{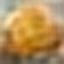}\\
\textbf{flamingo}  partridge      & \textbf{helicopter}  shotgun & \textbf{alligator}  crocodile & \textbf{bowl}  dish\\   
\includegraphics[scale=0.5]{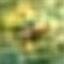}\includegraphics[scale=0.5]{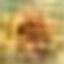}    &\includegraphics[scale=0.5]{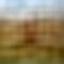}\includegraphics[scale=0.5]{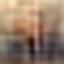} & \includegraphics[scale=0.5]{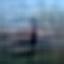}\includegraphics[scale=0.5]{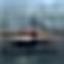}    &\includegraphics[scale=0.5]{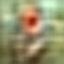}\includegraphics[scale=0.5]{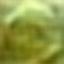}\\
\textbf{turtle} tortoise & \textbf{barn} cabinet & \textbf{sailboat}  boat & \textbf{emerald}  parsley\\  
\includegraphics[scale=0.5]{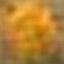}\includegraphics[scale=0.5]{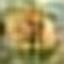} &\includegraphics[scale=0.5]{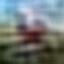}\includegraphics[scale=0.5]{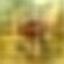} &\includegraphics[scale=0.5]{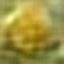}\includegraphics[scale=0.5]{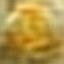} &\includegraphics[scale=0.5]{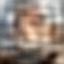}\includegraphics[scale=0.5]{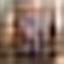}\\
\textbf{pumpkin} mandarin& \textbf{whale} bison	& \textbf{asparagus}  spinach & \textbf{thermometer}  marble\\
\end{tabular}
\caption{Proportion and examples of cases where subjects significantly
  preferred the dreamed concept image (left) or the confounder
  (right). Pairs of images depict dreamed and confounder
  concepts. Bold marks the dreamed concept that subjects were asked to
  pick the image for. E.g., subjects were presented with the bowl and
  dish images, were asked to decide which one contains a bowl, and
  preferred the image of the dish.}
\label{tab:exp4}
\end{center}
\end{table}

We next ran a follow-up experiment to find out to what extent the lack
of precision of our algorithm 
should be attributed
to noise in image generation from abstract visual features,
independently of the linguistic origin of the signal. For these
purposes, we replaced the visual feature vector produced by
cross-modal mapping with the ``gold-standard'' visual vector for each
dreamed/confounder concept (e.g., instead of mapping the
\emph{partridge} word vector onto visual space, we generated a
\emph{partridge} image  by inverting a \model{pool-5+exemplar}
vector directly extracted from a set of images labeled with this
word obtained from ImageNet). 
We repeated the Experiment 2 setup
using the images generated in this way. In this case, the number of
pairs for which no significant preference emerged was 75.4\%
(356/472), in 17.6\% (83/472) of the cases there was a significant
preference for the correct image, and in 7\% (33/472) for the
confounder. The results in this setting are better than when visual
features are derived from word representations, but not dramatically
so. Since feature inversion is an active area of research in computer
vision, we can thus expect that the quality of language-driven image
generation will greatly improve simply in virtue of general advances
in image generation methods.




\subsection{Experiment 3: Judging macro-categories of objects}

The previous experiments have shown that our language-driven image
generation system visualizes properties that are salient and relevant
enough to distinguish unrelated concepts (Experiment 1) but not
closely related ones (Experiment 2). The last experiment takes
high-level category structure explicitly into account in the design.

\paragraph{Experiment description}

We group the McRae categories into three macro-categories, namely
\textsc{animal} vs.\ \textsc{organic} vs.\ \textsc{man-made}, that are widely recognized in
cognitive science as fundamental and unambiguous
\cite{McRae:etal:2005}.  Participants are given a generated image and
are asked to pick the macro-category that best describes the object in
it.

\paragraph{Results} Again, the number of images for which
participants' preferences are not significant is high: 28\% of the
\textsc{organic} images, 47\% of the \textsc{man-made} images and 56\% of the \textsc{animal}
images.  However, when participants do show significant preference, in
the large majority of cases it is in favor of the correct
macro-category: this is so for 98\% of the \textsc{organic} images (70.5\% of
total), 90\% of the \textsc{man-made} images (48\% of total), and 59\% of the
\textsc{animal} ones (25.7\% of total). Table~\ref{tab:exp6} reports the
confusion matrix across the macro-categories. Confusions arise where
one would expect them: both \textsc{man-made} and \textsc{animal} images are more
often confused with \textsc{organic} things than with each other.



\begin{table}[t]
\begin{center}
\begin{tabular}{l| l l l | l l |l }
 \backslashbox{Gold}{Predicted}           &\textsc{man-made} &\textsc{organic} &\textsc{animal} & Pref. & No Pref. & Total\\\hline
  \textsc{man-made}  &128      &9      &5      &  142   &124     & 266\\
  \textsc{organic}   &0        &48      &1     & 49     &19      & 68\\
  \textsc{animal}    &9        &14      &33     & 56    &  72    & 128\\
\end{tabular}
\caption{Confusion matrix for experiment 3: rows report gold categories, columns report subjects' responses (the first 3 columns count cases with significant preferences for one macro-category only).}
\label{tab:exp6}
\end{center}
\end{table}

\begin{figure}[t]
\begin{tabular}{ccc}
\includegraphics[scale=0.27]{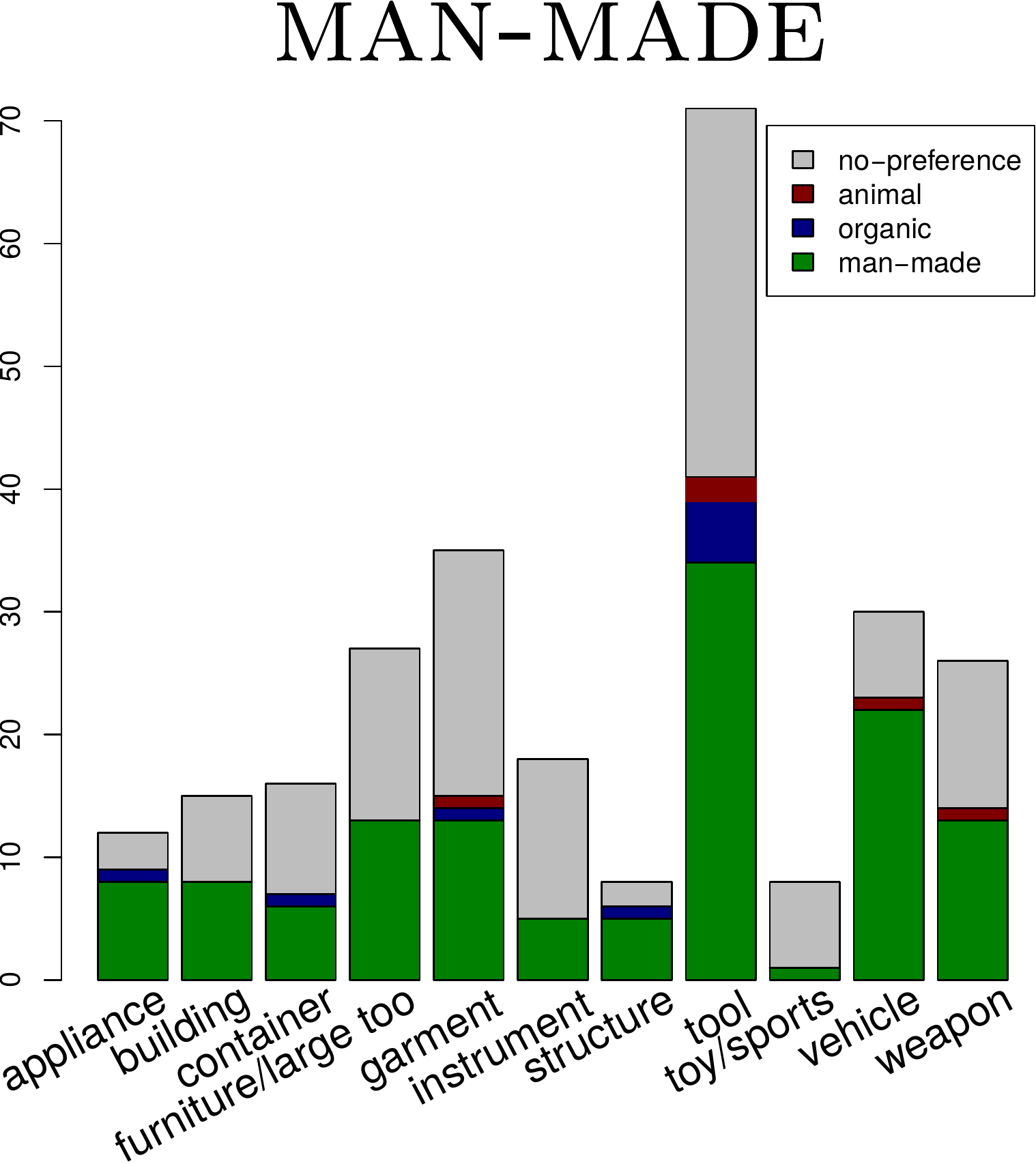} &
\includegraphics[scale=0.27]{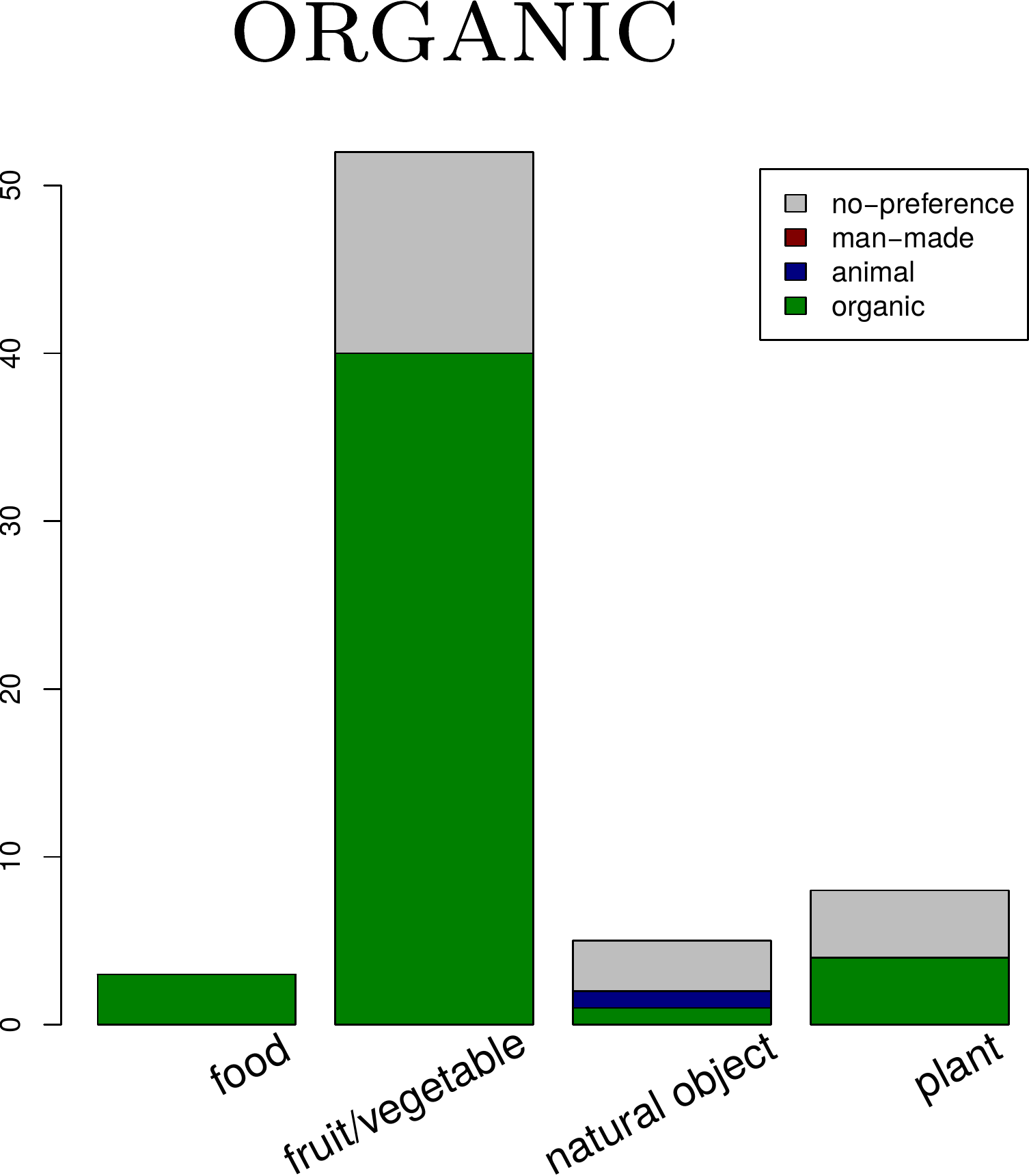} &
\includegraphics[scale=0.27]{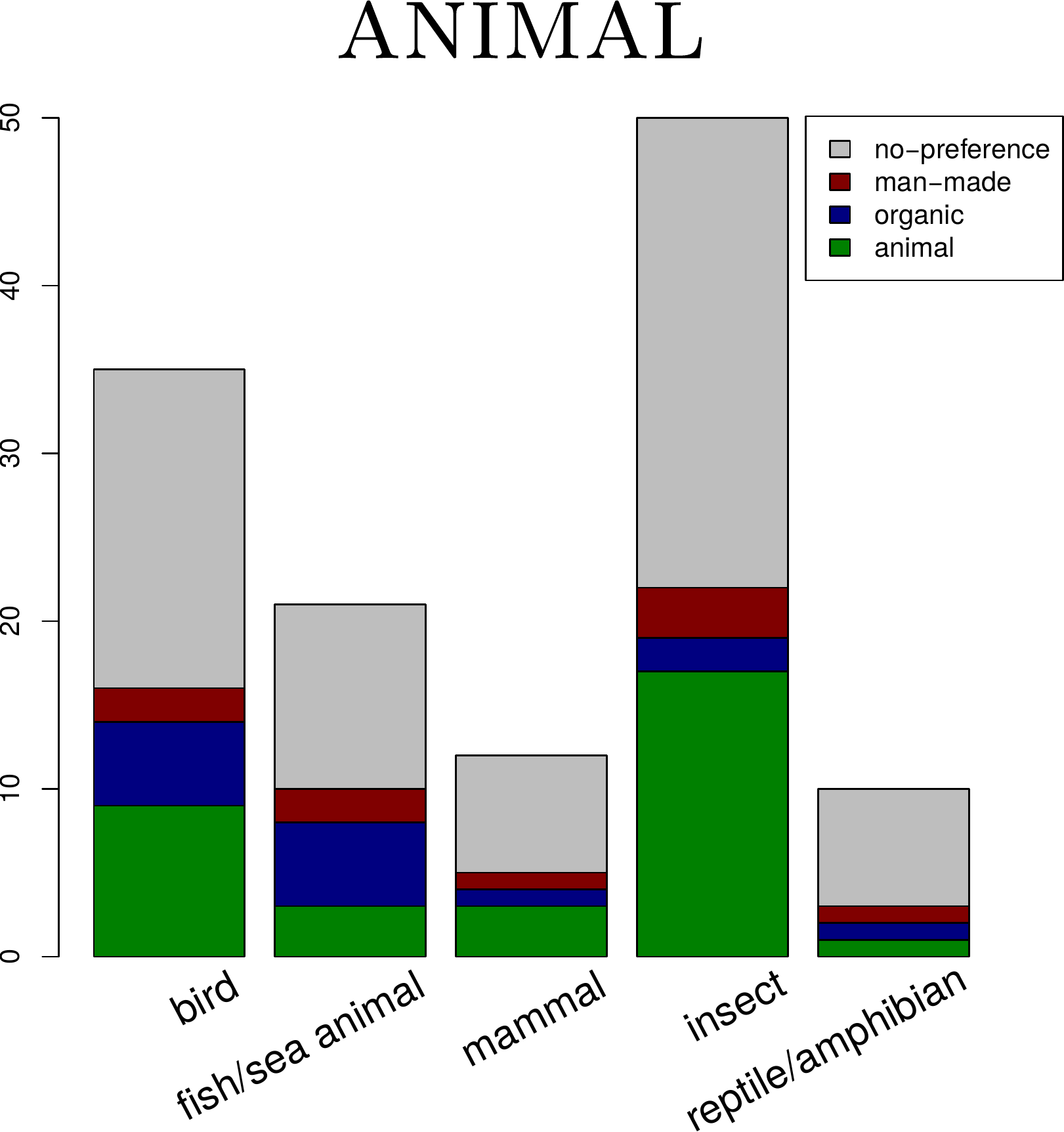}
\end{tabular}
\caption{Distribution of macro-category preferences across the gold
  concepts of the \textsc{man-made} (left), \textsc{organic} (middle) and \textsc{animal} (right)
  categories.}\label{fig:macro}
\end{figure}

Again, color (either of the object itself or of the environment) is
the leading property, distinguishing objects among the three
macro-categories.  As Figure~\ref{fig:summary} shows, orange, green
and a darker mixture of colors characterize \textsc{organic} things, \textsc{animal}s,
and \textsc{man-made} objects respectively.  Images that do not typically have
these colors are harder to be recognized.  For instance, the few
mistakes for \textsc{organic} images belong to the \emph{natural object}
category (e.g., rocks); all the other
categories within this macro-category are in the vast majority of the
cases judged correctly. In the \textsc{man-made}
macro-category (Figure~\ref{fig:macro}, left), the images of buildings are those more easily
recognizable; as one can see in Figure~\ref{fig:summary} those images
share the same pattern: two horizontal layers (land/dark and sky/blue)
with a vertical structure cutting across them (the building itself).
Similarly, vehicles display two layers with a small horizontal
structure crossing them, and they are almost always correctly
classified.  Finally, within the
\textsc{animal} macro-category (Figure~\ref{fig:macro}, right), birds and fish are more often misclassified
than other animals , with their typical environment probably playing a role in the confusion.

\section{Discussion}
\label{sec:discussion}

We introduced the new task of generating pictures visualizing the
semantic content of linguistic expressions as encoded in word embeddings, 
proposing more specifically a method we dubbed language-driven image generation.

The current system seems capable to visualize the typical color of
object classes and aspects of their characteristic
environment. Interestingly, vector-based word representations are
notoriously bad at capturing color \cite{Bruni:etal:2012a}, and we do
not expect them to be much better at characterizing environments, so
our results suggest that, already in its current form, our system
could also be used to \emph{enrich} word representations, by
highlighting aspects of concepts that are not salient in language but
are probably learned by similarity-based generalization from the
cross-modal mapping training examples. In this sense, language-driven
image generation is more than a simple word embedding evaluation
tool. At the same time, our system completely ignores visual
properties related to shape. Shapes are not often expressed by
linguistic means (although we all recognize the typical ``gestalt''
of, say, a mammal, it is very difficult to describe it in words), but
in the same way in which we can capture color and environment, better
visual representations or feature inversion methods might lead us in
the future to associate, by means of images, typical shapes to
shape-blind linguistic representations.

Currently we approach language-based image generation as a two-step
process. Inspired from recent work in caption generation that
conditions word production on visual vectors, we plan to explore an
end-to-end model that conditions the generation process on information
encoded in the word embeddings of the word/phrase that we wish to
produce an image for, building upon classic generative models 
of image generation~\cite{Salakhutdinov:Hinton:2009,Gregor:etal:2015}.

\bibliography{../../marco,../../angeliki}
\bibliographystyle{plain}
\begin{appendix}



\section{Answer Keys to Figure 1}

We provide the concept names of the word embeddings used to generate the 
images of Figure 1 (we provide again Figure 1 in this document to facilitate the
readers). Due to lack of space, we split the concept names into 3 tables, Table 1-3, where each table
provides the concept names of the word embeddings used to generate the \textsc{man-made}, \textsc{organic}
and \textsc{animal} images respectively.

\begin{figure*}[h]
\begin{centering}
\includegraphics[scale=0.58]{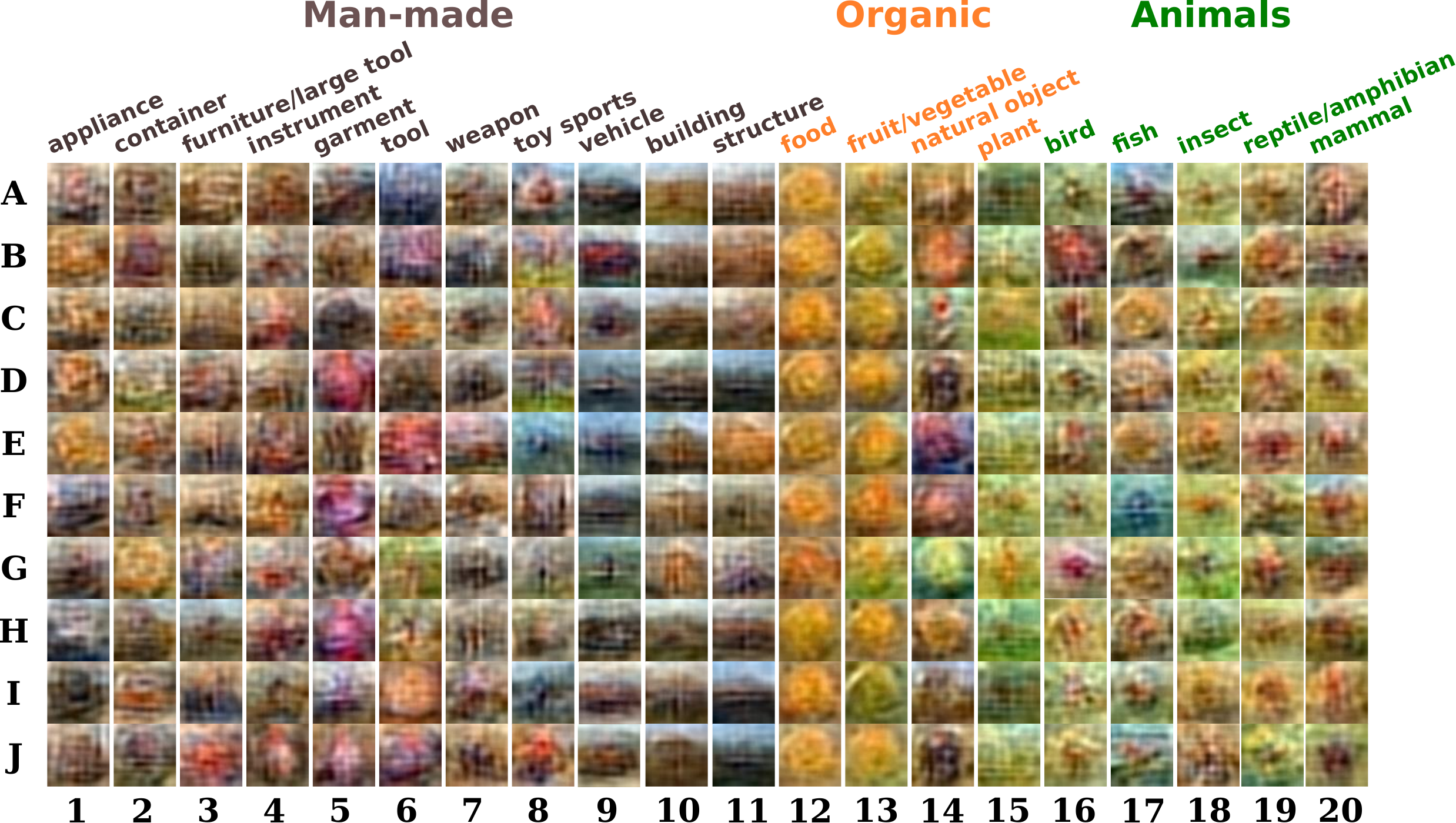}
\end{centering}
\caption{Generated images of 10 concepts per category for 20 basic
  categories, grouped my macro-category.\label{fig:summary}}
\end{figure*}

\begin{landscape}
\begin{table}
\begin{center}
\begin{tabular}{l||l|l|l|l|l|l|l|l|l|l|l}
&\textbf{1}&\textbf{2}&\textbf{3}&\textbf{4}&\textbf{5}&\textbf{6}&\textbf{7}&\textbf{8}&\textbf{9}&\textbf{10}&\textbf{11}\\\hline\hline
\textbf{A}&dishwasher&ashtray&bed&accordion&apron&anchor&axe&balloon&airplane&barn&apartment\\
\textbf{B}&freezer&bag&bench&bagpipe&armour&banner&baton&ball&ambulance&building&basement\\
\textbf{C}&fridge&barrel&bookcase&banjo&belt&blender&bayonet&doll&bike&bungalow&bedroom\\
\textbf{D}&microwave&basket&bureau&cello&blouse&bolts&bazooka&football&boat&cabin&bridge\\
\textbf{E}&oven&bathtub&cabinet&clarinet&boots&book&bomb&kite&buggy&cathedral&cellar\\
\textbf{F}&projector&bottle&cage&drum&bracelet&brick&bullet&marble&bus&chapel&elevator\\
\textbf{G}&radio&bowl&carpet&flute&buckle&broom&cannon&racquet&canoe&church&escalator\\
\textbf{H}&sink&box&catapult&guitar&camisole&brush&crossbow&rattle&cart&cottage&garage\\
\textbf{I}&stereo&bucket&chair&harmonica&cape&candle&dagger&skis&car&house&pier\\
\textbf{J}&stove&cup&sofa&harp&cloak&crayon&shotgun&toy&helicopter&hut&bridge\\
\end{tabular}
\caption{Concept names of word embeddings used to generate \textsc{man-made} images.}
\end{center}
\end{table}

\begin{table}[!htb]
 \begin{minipage}{.4\linewidth}
        \begin{tabular}{l||l|l|l|l}
&\textbf{12}&\textbf{13}&\textbf{14}&\textbf{15}\\\hline\hline
\textbf{A}&biscuit&apple&beehive&birch\\
\textbf{B}&bread&asparagus&bouquet&cedar\\
\textbf{C}&cake&avocado&emerald&dandelion\\
\textbf{D}&cheese&banana&muzzle&oak\\
\textbf{E}&pickle&beans&pearl&pine\\
\textbf{F}&pie&beets&rock&prune\\
\textbf{G}&raisin&blueberry&seaweed&vine\\
\textbf{H}&rice&broccoli&shell&willow\\
\textbf{I}&cake&cabbage&stone&birch\\
\textbf{J}&biscuit&cantaloupe&muzzle&pine\\
\end{tabular}
\caption{Concept names of word embeddings  used to generate \textsc{organic} images.}
\end{minipage}
\begin{minipage}{.1\linewidth}
\hspace{10ex}
\end{minipage}
 \begin{minipage}{.4\linewidth}
\centering
\begin{tabular}{l||l|l|l|l|l}
&\textbf{16}&\textbf{17}&\textbf{18}&\textbf{19}&\textbf{20}\\\hline\hline
\textbf{A}&blackbird&whale&grasshopper&alligator&bear\\
\textbf{B}&bluejay&octopus&hornet&crocodile&beaver\\
\textbf{C}&budgie&clam&moth&frog&bison\\
\textbf{D}&buzzard&cod&snail&iguana&buffalo\\
\textbf{E}&canary&crab&ant&python&bull\\
\textbf{F}&chickadee&dolphin&beetle&rattlesnake&calf\\
\textbf{G}&flamingo&eel&butterfly&salamander&camel\\
\textbf{H}&partridge&goldfish&caterpillar&toad&caribou\\
\textbf{I}&dove&guppy&cockroach&tortoise&cat\\
\textbf{J}&duck&mackerel&flea &cheetah&cheetah\\
\end{tabular}
\caption{Concept names of word embeddings used to generate \textsc{animal} images.}
\end{minipage}
\end{table}
\end{landscape}

\end{appendix}

\end{document}